\documentclass[11pt,a4paper]{article}

\usepackage[utf8]{inputenc}
\usepackage[T1]{fontenc}
\usepackage{lmodern}
\usepackage[margin=1in]{geometry}
\usepackage{graphicx}
\graphicspath{{../../gallery/}{./gallery/}{gallery/}}
\usepackage{booktabs}
\usepackage{amsmath,amssymb}
\usepackage{hyperref}
\usepackage{xcolor}
\usepackage{listings}
\usepackage{enumitem}
\usepackage{caption}
\usepackage{subcaption}
\usepackage{float}
\usepackage{algorithm}
\usepackage{algpseudocode}
\usepackage{natbib}
\usepackage{url}
\usepackage{fancyvrb}
\usepackage{tcolorbox}
\tcbuselibrary{breakable}

\lstdefinelanguage{json}{
    basicstyle=\ttfamily\small,
    morestring=[b]",
    literate=
     *{0}{{{\color{blue!70!black}0}}}{1}
      {1}{{{\color{blue!70!black}1}}}{1}
      {2}{{{\color{blue!70!black}2}}}{1}
      {3}{{{\color{blue!70!black}3}}}{1}
      {4}{{{\color{blue!70!black}4}}}{1}
      {5}{{{\color{blue!70!black}5}}}{1}
      {6}{{{\color{blue!70!black}6}}}{1}
      {7}{{{\color{blue!70!black}7}}}{1}
      {8}{{{\color{blue!70!black}8}}}{1}
      {9}{{{\color{blue!70!black}9}}}{1}
      {:}{{{\color{red!60!black}{:}}}}{1}
      {,}{{{\color{red!60!black}{,}}}}{1}
}

\hypersetup{
    colorlinks=true,
    linkcolor=blue!70!black,
    citecolor=blue!70!black,
    urlcolor=blue!70!black
}

\title{%
\textbf{Prober.ai: Gated Inquiry-Based Feedback via LLM-Constrained Personas for Argumentative Writing Development}
}
\author{
Ran Bi \quad Shiyao Wei$^{1}$ \quad Yuanyiyi Zhou$^{2}$ \\[4pt]
\small $^{1}$Florida State University \quad $^{2}$New York University
}
\date{July 2026}

\begin{document}

\maketitle

\begin{abstract}
The proliferation of large language models (LLMs) in educational settings has paradoxically undermined the cognitive processes they purport to support. Students increasingly outsource critical thinking to AI assistants that generate polished text on demand, resulting in measurable cognitive debt and diminished argumentative reasoning skills. We present \textsc{Prober.ai}, a web-based writing environment that inverts the conventional AI-tutoring paradigm: rather than generating or rewriting student text, the system constrains an LLM---OpenAI GPT-5.6, accessed through the Responses API with schema-constrained Structured Outputs---through persona-specific system prompts to produce only targeted, inquiry-based questions about argumentative weaknesses. A two-phase interaction architecture---\textit{Challenge} and \textit{Unlock}---implements a pedagogical friction mechanism whereby revision suggestions are gated behind an intelligent \textit{reflection gate}: the student's written defense is first assessed for substance and, if it is thin, coached rather than answered, so that cognitive effort provably precedes support. The system is grounded in Toulmin's argumentation theory, research on peer feedforward questioning mechanisms, and evidence on AI-induced cognitive debt, and it is fully instrumented for classroom study. A functional prototype was developed during OpenAI Build Week, with the application itself engineered using Codex (running on the GPT-5.6 Terra model) as the coding agent. We describe the system architecture, the methodology for constraining GPT-5.6 to pedagogically aligned JSON schemas via Structured Outputs, the design of the reflection gate, and implications for scalable, cognition-preserving AI integration in writing education.
\end{abstract}


\section{Introduction}
\label{sec:introduction}

The rapid integration of large language models (LLMs) into student writing workflows has created a pedagogical paradox. Tools such as ChatGPT, Gemini, and QuillBot offer immediate, fluent text generation and revision capabilities that students readily adopt. However, emerging neuroscience and behavioral evidence demonstrates that this convenience comes at a significant cognitive cost. Kosmyna et al.~\cite{kosmyna2025brain} measured EEG alpha-band activity during LLM-assisted essay writing and found statistically significant reductions in directed transfer function (dDTF) connectivity---a neural marker of active cognitive engagement---compared to both search-assisted and unassisted writing conditions. This phenomenon, characterized as \textit{cognitive outsourcing}, describes the systematic offloading of higher-order thinking processes to AI systems.

The consequences extend beyond neural metrics. Behaviorally, students who rely on AI-generated text produce essays that are superficially polished but structurally shallow: claims are asserted without adequate warrant, counterarguments are treated as perfunctory acknowledgments rather than genuine dialectical engagements, and the reasoning chains linking evidence to conclusions are frequently absent or circular~\cite{bi2026pedagogy}. Existing writing support tools exacerbate rather than address this problem. Grammar-focused platforms (Grammarly, QuillBot) operate at the surface level---correcting syntax, word choice, and tone---without engaging the argumentative structure of the text. General-purpose AI agents (ChatGPT, Gemini) provide direct answers, are systematically agreeable (``Sounds great! You have built a really strong, cohesive argument''), and never deliver the rigorous, challenging feedback that strengthens critical thinking.

This paper presents \textsc{Prober.ai}, a system designed to occupy a fundamentally different position in the AI-assisted writing landscape. The core design principle is that the AI should \textit{never write for the student}. Instead, the system constrains an LLM to function exclusively as a structured questioner---a ``devil's advocate'' that identifies logical weaknesses in the student's argumentative essay and poses targeted, open-ended questions that force the student to defend, clarify, and strengthen their own reasoning. Concrete revision suggestions are deliberately \textit{gated} behind a mandatory reflection step: the student must first articulate a written defense of their argument before the system unlocks specific, actionable feedback.

The technical contributions of this work are threefold:

\begin{enumerate}[leftmargin=*]
    \item \textbf{Persona-constrained LLM output.} We demonstrate a methodology for constraining a general-purpose LLM (GPT-5.6) to produce only pedagogically aligned, structured JSON output by combining carefully engineered system prompts with the OpenAI Responses API's \textit{Structured Outputs}---a strict, named JSON schema enforced by the model at decode time. This eliminates both the model's default tendency toward evaluative or generative responses and the parse-failure class inherent in prompt-only schema coaxing.

    \item \textbf{Reflection-gated feedback architecture.} We introduce a two-phase API design (\texttt{/challenge} $\rightarrow$ \texttt{/unlock}) in which the unlock step is guarded by an intelligent \textit{reflection gate}: a deterministic assessment of the student's defense (length, reasoning markers, question-specific relevance) that coaches low-effort reflections instead of unlocking. This implements pedagogical friction as a first-class architectural primitive, ensuring that genuine cognitive effort---not merely a keystroke---precedes the delivery of revision support.

    \item \textbf{Multi-persona questioning framework.} We operationalize two complementary critical personas---\textit{Reviewer \#2} (expert-level logical scrutiny) and \textit{Confused Reader} (novice-perspective clarity probing)---each producing distinct question taxonomies mapped to specific dimensions of argumentation quality.
\end{enumerate}

\section{Related Work}
\label{sec:related_work}

\subsection{AI-Assisted Feedback in Education}

Ba et al.~\cite{ba2025unraveling} conducted a systematic literature review of AI-assisted feedback in education, identifying that the majority of existing systems provide \textit{directive} feedback (explicit corrections and rewrites) rather than \textit{facilitative} feedback (questions and prompts that guide self-regulation). Their meta-analysis revealed that facilitative feedback mechanisms are more strongly associated with long-term learning gains, particularly in writing domains where the development of metacognitive awareness is a primary instructional goal. \textsc{Prober.ai} is explicitly designed as a facilitative feedback system, producing only questions in its initial interaction phase and gating directive suggestions behind student reflection.

\subsection{Argumentation Theory and Writing Assessment}

The question taxonomy employed by \textsc{Prober.ai} is grounded in Toulmin's model of argumentation~\cite{kinnear2022argumentation}, which decomposes arguments into claims, data (evidence), warrants (reasoning links), backing, qualifiers, and rebuttals. Kinnear et al.~\cite{kinnear2022argumentation} demonstrated how this framework can inform assessment validity in educational settings, providing a principled basis for identifying specific argumentative weaknesses. Our system operationalizes Toulmin's categories as distinct question modules: the \textit{claim question} targets the clarity and precision of the central thesis, the \textit{reasoning question} probes the warrant linking evidence to conclusion, the \textit{counterargument question} examines the depth of dialectical engagement, and the \textit{scope/implication question} addresses qualifiers and broader stakes.

\subsection{Peer Feedback and Question-Based Scaffolding}

Latifi et al.~\cite{latifi2021peer} investigated the distinction between peer feedback and peer feedforward in argumentative writing, finding that question-based feedforward---where reviewers pose questions rather than make evaluative statements---significantly enhanced both the quality of argumentation and the depth of the learning process compared to traditional feedback approaches. Their work demonstrated that questions activate different cognitive processes than statements: questions require the writer to generate rather than merely evaluate, shifting the locus of cognitive effort from the reviewer to the writer. \textsc{Prober.ai} extends this principle by replacing the human peer reviewer with a persona-constrained LLM, enabling on-demand, scalable question-based feedforward.

Noroozi et al.~\cite{noroozi2016relations} further established that scripted online peer feedback processes---where the feedback interaction is structured through predefined protocols---produce higher-quality argumentative essays than unscripted interactions. The structured JSON output schemas employed in \textsc{Prober.ai} serve an analogous function: they script the LLM's feedback behavior according to a pedagogically principled protocol, ensuring consistency and alignment with argumentation quality dimensions.

\subsection{Cognitive Outsourcing and AI Dependency}

Kosmyna et al.~\cite{kosmyna2025brain} provided the first neuroimaging evidence of cognitive debt accumulation during LLM-assisted writing. Their EEG study demonstrated that using ChatGPT for essay writing produced significantly lower alpha-band dDTF connectivity---indicating reduced active cognitive processing---compared to both search-engine-assisted and brain-only conditions. Critically, this effect persisted even when participants were instructed to use the AI as a ``thinking partner'' rather than a ghostwriter, suggesting that the mere availability of generated text suppresses independent reasoning. Gao et al.~\cite{gao2024students} extended this analysis to peer feedback contexts, finding that students' uptake of online peer feedback in argumentative essay writing was mediated by the cognitive effort required to process and integrate the feedback. These findings collectively motivate \textsc{Prober.ai}'s core design decision: by refusing to generate or rewrite text, the system eliminates the cognitive shortcut that enables outsourcing.

\subsection{Distinction from Existing Systems}

Table~\ref{tab:comparison} summarizes how \textsc{Prober.ai} differs from existing approaches. Unlike grammar-focused tools (Grammarly, QuillBot) that operate below the argumentative structure, and unlike general AI agents (ChatGPT, Gemini) that generate direct answers and systematically avoid harsh feedback, \textsc{Prober.ai} targets the logical structure of arguments and deliberately withholds solutions until the student has demonstrated reflective engagement.

\begin{table}[ht]
\centering
\caption{Comparison of \textsc{Prober.ai} with existing writing support paradigms.}
\label{tab:comparison}
\small
\begin{tabular}{@{}lccc@{}}
\toprule
\textbf{Feature} & \textbf{Writing Tools} & \textbf{General AI} & \textbf{Prober.ai} \\
 & \textit{(Grammarly, QuillBot)} & \textit{(ChatGPT, Gemini)} & \\
\midrule
Targets argument structure & \texttimes & Partial & \checkmark \\
Generates/rewrites text & \checkmark & \checkmark & \texttimes \\
Provides challenging feedback & \texttimes & \texttimes & \checkmark \\
Question-based interaction & \texttimes & \texttimes & \checkmark \\
Gates suggestions behind reflection & \texttimes & \texttimes & \checkmark \\
Structured output schema & \texttimes & \texttimes & \checkmark \\
Persona-driven feedback & \texttimes & \texttimes & \checkmark \\
\bottomrule
\end{tabular}
\end{table}

\section{System Architecture and Methodology}
\label{sec:architecture}

\subsection{Design Principles}

\textsc{Prober.ai} is architected around four core design principles derived from the theoretical foundations discussed in Section~\ref{sec:related_work}:

\begin{enumerate}[leftmargin=*]
    \item \textbf{Cognitive effort preservation.} The system must never reduce the cognitive load required for argumentation. Every interaction should increase or maintain the student's active reasoning engagement.

    \item \textbf{Question-based interaction.} The primary output modality is inquiry, not evaluation or generation. The system asks; it does not tell.

    \item \textbf{Necessary cognitive scaffolding.} While refusing to do the student's thinking, the system must provide sufficient structure to make the cognitive challenge productive rather than overwhelming.

    \item \textbf{Dual-perspective feedback.} Different argumentative weaknesses require different critical lenses. The system provides at least two complementary personas to address both logical rigor and communicative clarity.
\end{enumerate}

\subsection{High-Level Architecture}

\begin{figure}[ht]
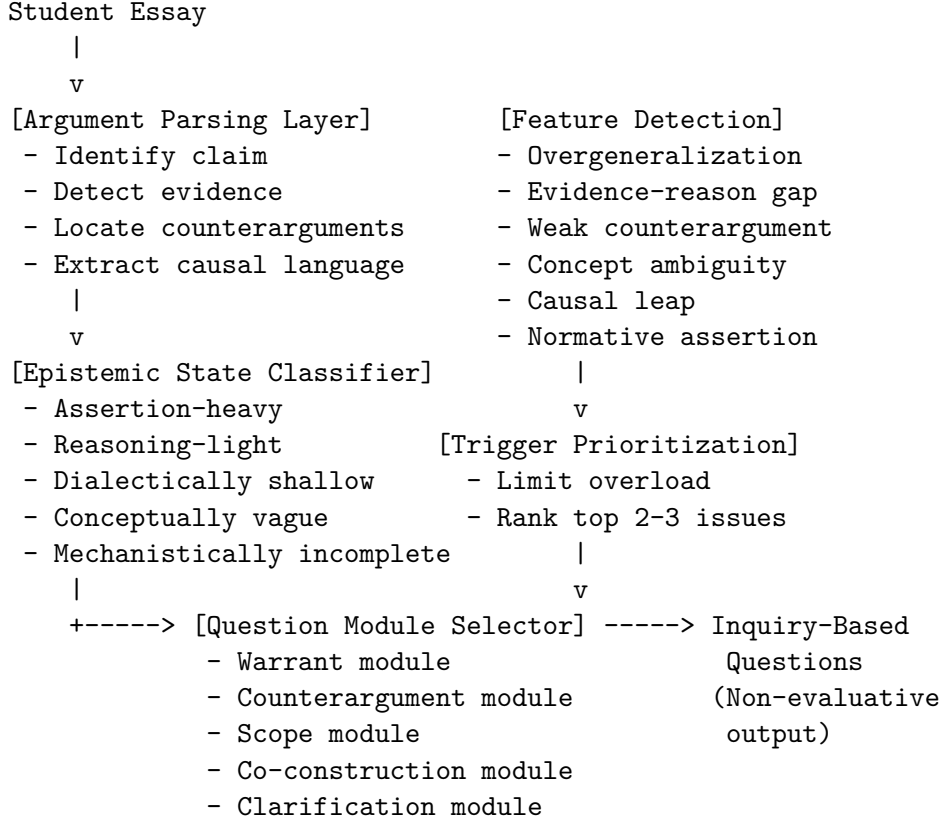

\centering
\begin{BVerbatim}
Student Essay
    |
    v
[Argument Parsing Layer]        [Feature Detection]
 - Identify claim               - Overgeneralization
 - Detect evidence              - Evidence-reason gap
 - Locate counterarguments      - Weak counterargument
 - Extract causal language      - Concept ambiguity
    |                           - Causal leap
    v                           - Normative assertion
[Epistemic State Classifier]         |
 - Assertion-heavy                   v
 - Reasoning-light          [Trigger Prioritization]
 - Dialectically shallow      - Limit overload
 - Conceptually vague         - Rank top 2-3 issues
 - Mechanistically incomplete        |
    |                                v
    +-----> [Question Module Selector] -----> Inquiry-Based
             - Warrant module                  Questions
             - Counterargument module         (Non-evaluative
             - Scope module                    output)
             - Co-construction module
             - Clarification module
\end{BVerbatim}
\caption{Conceptual processing pipeline of \textsc{Prober.ai}. The LLM performs argument parsing, feature detection, epistemic state classification, trigger prioritization, and question module selection as internal reasoning steps. Only the final inquiry-based questions are surfaced to the student.}
\label{fig:architecture}
\end{figure}

The system follows a client-server architecture with a clear separation between the writing environment (frontend) and the AI reasoning pipeline (backend). Figure~\ref{fig:architecture} illustrates the overall system structure.

\begin{figure}[t]
\centering
\includegraphics[width=\textwidth,keepaspectratio]{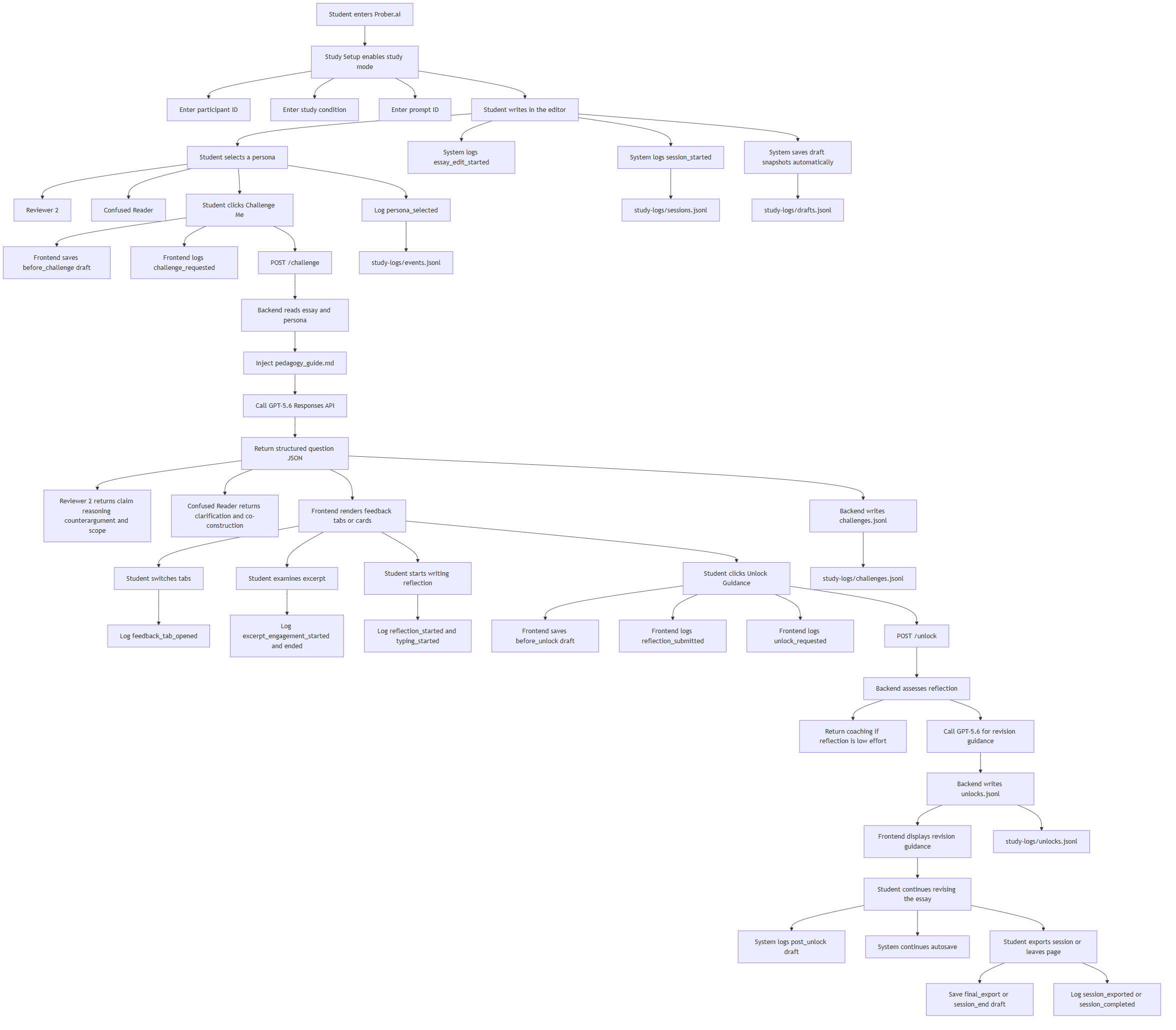}
\caption{System function and data flow. The student selects a persona and submits an essay; the backend injects the pedagogy guide and calls GPT-5.6 through the Responses API, returning schema-constrained questions. After the student writes a reflection, the \texttt{/unlock} endpoint assesses it and either returns a coaching nudge or calls GPT-5.6 for revision guidance. Every stage is written to the research pipeline (Section~\ref{sec:instrumentation}).}
\label{fig:system_flow}
\end{figure}

While Figure~\ref{fig:architecture} depicts the model's \emph{internal} reasoning, Figure~\ref{fig:system_flow} shows the concrete function and data flow of the deployed system: persona selection, the schema-constrained GPT-5.6 call on the Responses API, the reflection gate, and the research-logging pipeline that records every stage.

\subsection{The Challenge--Defend--Improve Loop}

\begin{figure}[ht]
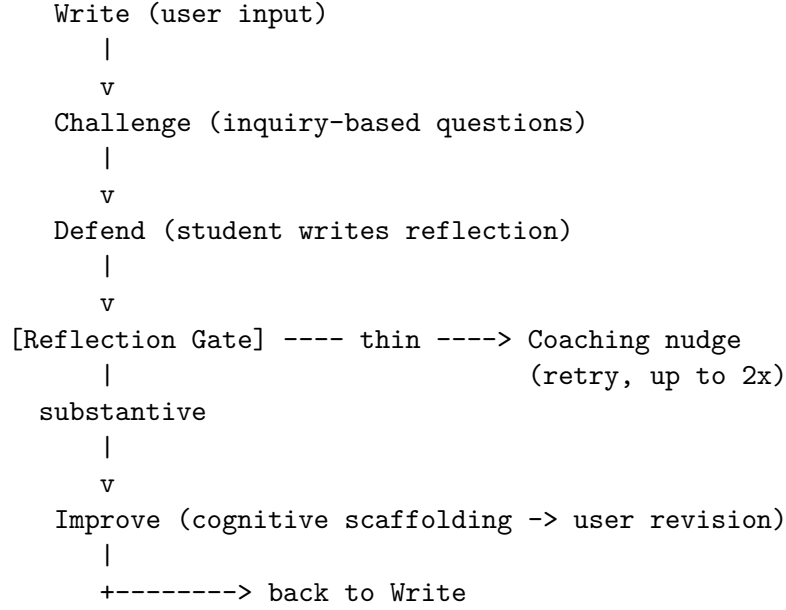

\centering
\begin{BVerbatim}
      Write (user input)
         |
         v
      Challenge (inquiry-based questions)
         |
         v
      Defend (student writes reflection)
         |
         v
   [Reflection Gate] ---- thin ----> Coaching nudge
         |                           (retry, up to 2x)
     substantive
         |
         v
      Improve (cognitive scaffolding -> user revision)
         |
         +--------> back to Write
\end{BVerbatim}
\caption{The Write--Challenge--Defend--Improve cycle. The reflection gate sits between Defend and Improve: a thin defense is coached and returned for another attempt, while a substantive defense unlocks a suggestion. The loop is designed so that cognitive effort always precedes the delivery of suggestions, ensuring the student remains the primary agent of revision.}
\label{fig:user_loop}
\end{figure}

The user interaction follows a cyclical four-phase model (Figure~\ref{fig:user_loop}):

\begin{enumerate}[leftmargin=*]
    \item \textbf{Write.} The student composes or pastes an argumentative essay into the Quill-based rich text editor.

    \item \textbf{Challenge.} The student selects a critical persona and submits their essay. The system returns structured, inquiry-based questions targeting specific argumentative dimensions. No evaluative language or revision suggestions are provided at this stage.

    \item \textbf{Defend.} For each question, the student must write a reflective defense articulating how they would address the identified weakness. This reflection step is the system's primary pedagogical friction mechanism, and it is not accepted uncritically: an intelligent \textit{reflection gate} (Section~\ref{sec:reflection_gate}) assesses whether the defense substantively engages the question before any suggestion is released.

    \item \textbf{Improve.} Once the defense clears the reflection gate, the student ``unlocks'' a concrete revision suggestion and a writing tip that build on the student's own reasoning. The student then incorporates these into their draft and may re-enter the loop with the revised text.
\end{enumerate}

\subsection{The Reflection Gate}
\label{sec:reflection_gate}

The reflection gate is the mechanism that makes pedagogical friction \emph{enforceable} rather than cosmetic. In the earlier prototype, the unlock step required only a non-empty defense string---typing a single character released the suggestion. The current system replaces that turnstile with a two-layer gate.

\paragraph{Layer 1: deterministic assessment.} Before any model call, the server scores the defense with a fast, free heuristic (\texttt{assessReflectionQuality}). It checks for (i)~filler or non-answers (e.g., ``idk,'' ``not sure''), (ii)~a minimum length floor (18 words), (iii)~the presence of reasoning markers (\textit{because}, \textit{therefore}, \textit{this shows}, \textit{however}, \ldots), and (iv)~topical relevance, measured as the content-word overlap between the defense and the specific question and excerpt being probed. These produce a 0--4 \texttt{reflectionScore} and a set of machine-readable \texttt{reasonCodes} (\texttt{too\_short}, \texttt{missing\_reasoning\_marker}, \texttt{not\_specific\_to\_question}, \texttt{low\_effort}).

\paragraph{Coaching instead of gatekeeping.} A defense that fails the assessment does not unlock a suggestion; instead the gate returns a targeted coaching nudge---for example, \textit{``Add the because: explain the reasoning link between your claim, the evidence, and the conclusion you want the reader to accept''}---selected from the specific reason the defense fell short. Crucially, the gate never traps an earnest student: after two coached attempts, a genuine effort (non-filler, at least 10 words) is allowed through to a more scaffolded hint. This operationalizes \textit{productive struggle}---the reflection itself becomes a second learning moment rather than an obstacle to game.

\paragraph{Layer 2: model-side gate.} Only when the deterministic layer is satisfied does the system call GPT-5.6, and the model is itself constrained (via the unlock schema) to return a \texttt{status} of either \texttt{coaching} or \texttt{unlocked}: it may still withhold the suggestion and nudge if the defense engages the wrong point. The accumulating corpus of \{question, defense, score, reason codes, revision\} is captured by the research pipeline (Section~\ref{sec:instrumentation}) and is the raw material for a future data-driven reflection-quality model.

\subsection{Persona System}

Two complementary personas are implemented, each addressing distinct dimensions of argumentative quality:

\subsubsection{Reviewer \#2: The Logical Assassin}

This persona emulates an expert-level academic peer reviewer with deep domain expertise. Its system prompt constrains the LLM to:

\begin{itemize}[leftmargin=*]
    \item Ignore prose, grammar, and flow; focus strictly on structural integrity of the argument.
    \item Identify logical ``black holes'' and theoretical flaws.
    \item Adopt a cold, clinical, intellectually demanding tone.
    \item Produce exactly four questions mapped to Toulmin's argumentation dimensions:
    \begin{enumerate}
        \item \texttt{claim\_question}: Probes the clarity and precision of the central thesis or key sub-claims.
        \item \texttt{reasoning\_question}: Examines the warrant linking evidence to conclusion.
        \item \texttt{counterargument\_question}: Invites deeper engagement with opposing views.
        \item \texttt{scope\_or\_implication\_question}: Raises issues of scope, boundary conditions, or larger implications.
    \end{enumerate}
\end{itemize}

\subsubsection{Confused Reader: The Frustrated Novice}

This persona emulates an intelligent outsider who experiences the ``curse of knowledge''---the gap between what the author assumes the reader knows and what the reader actually understands. Its constraints include:

\begin{itemize}[leftmargin=*]
    \item Identify where cognitive load becomes excessive: jargon, undefined concepts, or explanatory leaps.
    \item Pinpoint exactly where the reader felt ``lost.''
    \item Produce exactly two questions:
    \begin{enumerate}
        \item \texttt{clarification\_question}: Directly asks the writer to clarify a confusing term, leap in logic, or missing definition.
        \item \texttt{co\_construction\_question}: Invites the writer to brainstorm alternative possibilities or explanations collaboratively.
    \end{enumerate}
\end{itemize}

The dual-persona design ensures that students receive feedback on both the \textit{logical rigor} (Reviewer \#2) and the \textit{communicative clarity} (Confused Reader) of their arguments, targeting the two most critical and frequently independent dimensions of argumentative writing quality.

\subsection{LLM Constraint Methodology}

A central technical challenge is constraining a general-purpose LLM---whose default behavior includes evaluation, rewriting, and agreeableness---to produce \textit{only} structured questions without evaluative language. Our approach combines three constraint mechanisms:

\subsubsection{System Prompt Engineering}

The system prompt explicitly specifies what the model must \textit{not} do, using negative constraints that override default LLM behaviors:

\begin{itemize}[leftmargin=*]
    \item ``Do NOT rewrite the student's text.''
    \item ``Do NOT evaluate with words like `unclear', `weak', or `insufficient'.''
    \item ``Avoid yes/no questions.''
    \item ``Avoid leading the student toward a specific answer.''
    \item ``Avoid paraphrasing large chunks of the student's text.''
\end{itemize}

\subsubsection{Internal Reasoning Protocol}

The prompt specifies a multi-step internal reasoning chain that the model must execute \textit{without outputting}:

\begin{enumerate}
    \item \textbf{Argument segmentation:} Internally decompose the essay into claims, sub-claims, evidence instances, counterarguments, rebuttals, conclusions, definitions, and normative recommendations.

    \item \textbf{Issue detection:} Identify potential weaknesses from a diagnostic trigger list: overgeneralization, evidence--reasoning gaps, weak counterarguments, conceptual ambiguity, causal leaps, normative claims without value frameworks, and lack of implications.

    \item \textbf{Epistemic state classification:} Infer a holistic characterization of the essay's argumentative state (e.g., assertion-heavy, reasoning-light, dialectically shallow, conceptually vague, mechanistically incomplete, normatively under-justified).

    \item \textbf{Trigger prioritization:} Rank the top 2--3 issues to avoid cognitive overload on the student.
\end{enumerate}

Under the Responses API, this internal deliberation is further supported by GPT-5.6's native \texttt{reasoning.effort} control, which is set to \texttt{medium} for the analysis-heavy \texttt{/challenge} step and \texttt{low} for the lighter \texttt{/unlock} step, balancing analytical depth against interactive latency.

\subsubsection{Structured Outputs}

Rather than coaxing the model toward a schema through prose and then parsing whatever it returns, \textsc{Prober.ai} uses the OpenAI Responses API's \textit{Structured Outputs}: each request supplies a named JSON schema with \texttt{strict:\ true}, and the model is constrained at decode time to emit only conformant JSON. This guarantees a valid, fully-typed object on every call and eliminates the parse-failure class that prompt-only schema coaxing incurs. Listing~\ref{lst:schema} shows the schema enforced for the Reviewer~\#2 persona.

\begin{lstlisting}[language=json, caption={Strict JSON schema supplied to Structured Outputs for the Reviewer \#2 persona. Every field is required; excerpt fields are nullable so the model can decline to anchor a question.}, label={lst:schema}]
{
  "type": "object",
  "additionalProperties": false,
  "required": [
    "claim_question", "reasoning_question",
    "counterargument_question", "scope_or_implication_question",
    "claim_excerpt", "reasoning_excerpt",
    "counterargument_excerpt", "scope_or_implication_excerpt"
  ],
  "properties": {
    "claim_question":                { "type": "string" },
    "reasoning_question":            { "type": "string" },
    "counterargument_question":      { "type": "string" },
    "scope_or_implication_question": { "type": "string" },
    "claim_excerpt":                 { "type": ["string", "null"] },
    "reasoning_excerpt":             { "type": ["string", "null"] },
    "counterargument_excerpt":       { "type": ["string", "null"] },
    "scope_or_implication_excerpt":  { "type": ["string", "null"] }
  }
}
\end{lstlisting}

Each question field is additionally constrained by the system prompt to stand alone (no bullet lists), not exceed 2--3 sentences, and avoid concrete suggestions or content. The nullable excerpt fields let the model anchor a question to a specific passage; the server then verifies that every returned excerpt is an exact substring of the essay and silently drops any that is not, so hallucinated quotations can never drive the frontend's contextual highlighting.

\subsection{Pedagogy Guide Integration}

An external knowledge base (\texttt{pedagogy\_guide.md}, 128 lines) is loaded at server startup and injected into every \texttt{/challenge} prompt as internal context. This document codifies the question module taxonomy (Warrant/Reasoning, Counterargument, Scope/Overgeneralization, Normative Foundation, Conceptual Precision, Implication \& Stakes, Co-Construction, and Clarification modules) along with example question templates. The guide is prefaced with the instruction ``for your internal use only, do not quote or mention it explicitly,'' ensuring that the pedagogical framework shapes the model's questioning behavior without being surfaced to the student.

\section{Implementation Details}
\label{sec:implementation}

\subsection{Development Methodology}
\label{sec:devmethod}

Consistent with the venue, \textsc{Prober.ai} used OpenAI on two levels---to \textit{build} the system and to \textit{run} it. The application was engineered with \textbf{Codex}, OpenAI's agentic coding tool (running on the GPT-5.6 Terra model), working from plain-language instructions. Codex read the existing repository and the product-improvement notes, implemented the schema-constrained \texttt{/challenge} and \texttt{/unlock} handlers on the Responses API, built the reflection gate and its coaching heuristics, rewired the front end for the coached-reflection state, and updated the setup documentation, sample and demo content, dependency manifest, and validation scripts. Product identity---an AI that asks hard questions and waits for the student's own reasoning---and the decision to invest in the reflection gate rather than a broad redesign were kept in human hands. The runtime behavior described in the remainder of this section is the result of that collaboration.

\subsection{Technology Stack}

The prototype was implemented as a full-stack web application with the following components:

\begin{itemize}[leftmargin=*]
    \item \textbf{Backend:} Node.js with Express.js (v4.21.2), serving as both a static file server and an API gateway to the model. The server logic lives in a single file (\texttt{server.js}, roughly 700 lines), reflecting a deliberately lean prototype.

    \item \textbf{LLM Integration:} OpenAI's \texttt{openai} SDK (v6.48.0) calling the \texttt{gpt-5.6} model through the \textbf{Responses API}. Every call requests Structured Outputs (a strict, named JSON schema), sets a \texttt{reasoning.effort} tier (\texttt{medium} for \texttt{/challenge}, \texttt{low} for \texttt{/unlock}), sets an output verbosity level, disables server-side response storage (\texttt{store: false}), and attaches a stable, privacy-preserving \texttt{safety\_identifier} derived by hashing the study/session context. A 45-second request timeout and automatic retries guard against transient failures.

    \item \textbf{Frontend:} Vanilla HTML/CSS/JavaScript with no framework dependencies. The rich text editor is provided by Quill.js (Snow theme), offering a familiar word-processor experience with programmatic access to text content and formatting via the Quill Delta API.

    \item \textbf{Environment Management:} \texttt{dotenv} (v16.4.7) for the server-side \texttt{OPENAI\_API\_KEY}, with a client-side override: users may paste their own OpenAI key on a dedicated login page, where it is held only in the browser's \texttt{sessionStorage} and forwarded per request via an \texttt{x-openai-api-key} header (used only by \texttt{/challenge} and \texttt{/unlock}).

    \item \textbf{Deployment:} Vercel serverless platform via \texttt{@vercel/node}, with a \texttt{vercel.json} configuration routing all requests through the Express application. The deployment configuration explicitly includes static assets, sample essays, and the pedagogy guide in the serverless bundle.
\end{itemize}

\subsection{API Endpoints}

\subsubsection{\texttt{POST /challenge}}

This endpoint implements the questioning phase. It accepts:

\begin{itemize}[leftmargin=*]
    \item \texttt{essay} (string, required): The student's essay text (capped at 20{,}000 characters; oversized payloads are rejected with a 413 before any model call).
    \item \texttt{persona} (string, optional, default: \texttt{reviewer2}): The selected critical persona (\texttt{reviewer2} or \texttt{confusedReader}).
    \item \texttt{study} (object, optional): Participant, session, condition, and prompt identifiers used only for research logging (Section~\ref{sec:instrumentation}).
\end{itemize}

An optional \texttt{x-openai-api-key} request header supplies a user's own key, taking priority over the server's \texttt{OPENAI\_API\_KEY}. The endpoint constructs a composite prompt from (1)~the persona-specific system prompt, (2)~global constraints prohibiting evaluative and generative output, (3)~the full pedagogy guide as internal context, and (4)~the student's essay; the output schema itself is supplied out-of-band through the Responses API's Structured Outputs parameter, the reasoning effort is set to \texttt{medium}, and the output is capped at 1{,}200 tokens. Because Structured Outputs guarantees a schema-valid object, the response is consumed directly; a defensive parser that unwraps \texttt{```json} fences remains only as a belt-and-suspenders fallback. Each returned excerpt is validated as an exact substring of the essay before use. For the Confused Reader persona, backwards-compatible generic fields (\texttt{claim\_question}, \texttt{reasoning\_question}) are populated from the specialized \texttt{clarification\_question} and \texttt{co\_construction\_question} fields to simplify frontend rendering logic.

\subsubsection{\texttt{POST /unlock}}

This endpoint implements the reflection-gated suggestion phase (Section~\ref{sec:reflection_gate}). It accepts:

\begin{itemize}[leftmargin=*]
    \item \texttt{essay} (string, required): The full essay text for context.
    \item \texttt{label} (string): The category of the challenge (e.g., ``CLAIM'', ``REASONING'').
    \item \texttt{excerpt} (string, optional): The specific passage the question referenced.
    \item \texttt{question} (string, required): The original challenge question.
    \item \texttt{userDefense} (string, required): The student's written reflection/defense.
    \item \texttt{unlockAttemptCount} (integer, optional): How many times the student has attempted this defense, used by the gate to decide when an earnest but struggling student should be released to a scaffolded hint.
\end{itemize}

The endpoint first runs the deterministic reflection assessment. If the defense is not yet ready, it returns a \textit{coaching} payload immediately, \textit{without any model call}. Only a defense that clears the assessment reaches the model, which uses a \textit{different} persona---a ``helpful writing tutor'' that protects student thinking---rather than the adversarial questioning persona. The prompt provides the original question, the relevant excerpt, and the student's defense, then asks the model either to nudge once more (if the defense engages the wrong point) or to give concrete revision moves that build on the student's own reasoning, explicitly forbidding a full ready-to-submit replacement paragraph (suggestion under 130 words, tip under 25 words). This call runs at \texttt{low} reasoning effort with a 700-token cap. The response conforms to the schema \texttt{\{status, coachingMessage, reflectionScore, reasonCodes, suggestion, tip\}}, where \texttt{status} is \texttt{coaching} or \texttt{unlocked}.

This architectural separation ensures that the adversarial questioning phase and the supportive suggestion phase are handled by distinct prompt configurations, preventing persona contamination, while the two-layer gate (deterministic assessment followed by the model-side \texttt{status}) ensures that support is released only after genuine reflection.

\subsection{Frontend Architecture}

\subsubsection{Main Application (\texttt{/app})}

The main application (\texttt{index.html} + \texttt{script.js}, roughly 1{,}600 lines) manages the following state:

\begin{itemize}[leftmargin=*]
    \item \texttt{currentPersona}: Drives persona selection and determines which JSON fields to render.
    \item \texttt{useTabsView}: Toggles between a tabbed interface (one question at a time) and a card-based layout (all questions visible simultaneously).
    \item \texttt{sessionLog}: Accumulates the full interaction history (questions, defenses, suggestions) for session export.
    \item \texttt{totalChallenges} / \texttt{unlockedCount}: Track progress through the challenge--defend cycle with a visual progress indicator.
\end{itemize}

A key interaction feature is \textit{excerpt highlighting}: when the user hovers over a feedback card, the system performs a substring search of the excerpt text within the Quill editor's content and applies a semi-transparent yellow background (\texttt{rgba(250, 204, 21, 0.4)}) via \texttt{quill.formatText()}, visually linking the question to the relevant passage in real time. When the reflection gate returns a coaching payload, the frontend renders the nudge inline---together with the number of coached attempts remaining---keeping the student in the same reflection context rather than surfacing it as an error.

\subsubsection{Demo Mode (\texttt{/demo})}

A self-contained demo mode (\texttt{demo.html} + \texttt{demo.js}, 683 lines) provides a fully functional preview of the system without requiring an API key or server connectivity. It includes:

\begin{itemize}[leftmargin=*]
    \item A pre-loaded sample essay (a K--12 argumentative essay about driverless cars, exhibiting multiple common weaknesses: overgeneralization, causal leaps, weak counterarguments, and scope issues).
    \item Pre-baked feedback for both personas with carefully crafted questions and excerpt anchors.
    \item A local re-implementation of the reflection gate: a low-effort defense is \textit{coached} (with attempt tracking and a final scaffolded hint), exactly mirroring the server behavior, so the gate---the product's centerpiece---is demonstrable offline.
    \item Pre-baked unlock suggestions for all six question types, enabling the full Write--Challenge--Defend--Improve loop without any API calls.
\end{itemize}

\subsubsection{Session Export}

The system generates a print-ready HTML document containing the complete interaction log: the essay excerpt, each challenge question with its associated excerpt, the student's reflective defense, the AI's revision suggestion, and writing tips. This export serves both as a learning artifact for the student and as potential research data for educators.

\subsection{Research Instrumentation}
\label{sec:instrumentation}

Because \textsc{Prober.ai} is intended as the basis for empirical study, the system ships with a research-logging pipeline that captures the full interaction without altering the student experience. An optional \textit{Study Setup} mode records a participant identifier, session identifier, study condition, and prompt identifier, together with an explicit consent flag. Three dedicated endpoints (\texttt{/study/session}, \texttt{/study/event}, and \texttt{/study/draft}), alongside the \texttt{/challenge} and \texttt{/unlock} handlers, append newline-delimited JSON to six logs:

\begin{itemize}[leftmargin=*,noitemsep]
    \item \texttt{sessions.jsonl} --- session metadata and the consent flag;
    \item \texttt{events.jsonl} --- the timestamped interaction timeline;
    \item \texttt{drafts.jsonl} --- autosaved, version-numbered essay snapshots;
    \item \texttt{challenges.jsonl} --- each persona challenge and the questions returned;
    \item \texttt{unlocks.jsonl} --- each cleared defense and the unlocked guidance;
    \item \texttt{reflection-coaching.jsonl} --- reflections the gate coached rather than unlocked.
\end{itemize}

Fine-grained events---persona selection, excerpt engagement, reflection typing, unlock requests, coaching, and export---are timestamped, so a researcher can reconstruct each student's revision trajectory and, crucially, the \{question, defense, score, revision\} tuples the reflection gate produces. Each model call carries a \texttt{safety\_identifier} formed by hashing the study or session context, giving a stable but privacy-preserving per-participant signal. Appendix~\ref{app:logschema} details the resulting log schema. As noted in Section~\ref{sec:discussion}, this pipeline currently writes to local disk and must move to a durable datastore before production use.

\section{Preliminary Evaluation and Proof of Concept}
\label{sec:evaluation}

\subsection{Development Context}

\textsc{Prober.ai} was developed during OpenAI Build Week by a three-person team spanning development, learning-sciences research, and UX/UI design. As described in Section~\ref{sec:devmethod}, the application was engineered with Codex on the GPT-5.6 Terra model, while the team retained authority over the product's identity and over the decision to make the reflection gate the centerpiece of the build.

\subsection{Functional Validation}

The prototype was validated through iterative testing with sample essays spanning multiple genres and quality levels. Key observations include:

\begin{itemize}[leftmargin=*]
    \item \textbf{Schema compliance.} Because Structured Outputs constrains GPT-5.6 to a strict schema at decode time, every \texttt{/challenge} and \texttt{/unlock} call returned a valid, fully-typed object during development. The parse-failure and fallback-recovery class that a prompt-only schema incurs was eliminated by construction rather than mitigated after the fact.

    \item \textbf{Question quality.} Questions generated by the Reviewer \#2 persona consistently targeted genuine argumentative weaknesses rather than surface-level issues. The model reliably distinguished between claim-level, reasoning-level, counterargument-level, and scope-level concerns, producing questions that were categorically distinct across the four dimensions.

    \item \textbf{Persona differentiation.} The Confused Reader persona produced qualitatively different questions from the Reviewer \#2 persona on the same essay: where Reviewer \#2 challenged logical structure, the Confused Reader identified jargon, unexplained concepts, and explanatory gaps---confirming that the persona constraint mechanism effectively alters the model's analytical focus.

    \item \textbf{Gating effectiveness.} The reflection gate reliably distinguished low-effort defenses from substantive ones: filler and one-word answers were coached rather than answered, while defenses that contained genuine reasoning engaging the specific question unlocked guidance. Because the first assessment layer is deterministic, this behavior is reproducible and testable without a model call.
\end{itemize}

\subsection{Latency Considerations}

GPT-5.6's per-call latency is tuned through the Responses API's reasoning-effort tiers: \texttt{/challenge}, which performs the deeper argument analysis, runs at \texttt{medium} effort, while \texttt{/unlock} runs at \texttt{low} effort for faster, more interactive feedback. The reflection gate contributes a further benefit: low-effort defenses are handled entirely by the deterministic assessment layer and return \textit{instantly}, with no model round-trip at all. A 45-second request timeout with automatic retries bounds worst-case latency; server-sent streaming (Section~\ref{sec:discussion}) would further reduce perceived latency in a production deployment.

\subsection{Target Audience Validation}

The system was designed for two primary use cases identified through the team's pedagogical research:

\begin{enumerate}[leftmargin=*]
    \item \textbf{Regular English language arts learning:} K--12 students developing argumentative writing skills in standard ELA curricula. The demo essay (a middle-school-level argumentative essay about driverless cars) was specifically selected to represent this population's typical writing quality and common weaknesses.

    \item \textbf{Exam-based argumentative skill improvement:} Students preparing for standardized assessments (AP English Language, GRE Analytical Writing) where argumentative structure, logical coherence, and engagement with counterarguments are explicitly evaluated.
\end{enumerate}

\section{Discussion and Future Work}
\label{sec:discussion}

\subsection{Limitations}

Several limitations of the current prototype must be acknowledged:

\begin{enumerate}[leftmargin=*]
    \item \textbf{Absence of controlled evaluation.} The system has not been evaluated in a controlled experimental setting with student participants. Claims about cognitive engagement preservation, argumentative writing improvement, and learning outcomes remain theoretical, grounded in prior literature rather than empirical validation specific to this system. The system is now fully instrumented for such a study (Section~\ref{sec:instrumentation}), but no controlled data has yet been collected.

    \item \textbf{LLM output variability.} While Structured Outputs guarantees the \textit{form} of the model's output, the pedagogical \textit{quality} and alignment of individual questions still varies across invocations. The system lacks a post-generation quality filter or rubric-based validation layer that could reject or regenerate a weak question.

    \item \textbf{Single-turn interaction per question.} The reflection gate adds up to two coaching turns on the \textit{defense}, but the architecture still does not support open-ended multi-turn dialogue around a single question. If an unlocked suggestion is misaligned, there is no mechanism for iteratively refining the defense--suggestion exchange.

    \item \textbf{Genre specificity.} The system's question modules and diagnostic triggers are optimized for argumentative/persuasive essays. Extending to other genres (narrative, expository, analytical) would require redesigning the argumentation parsing heuristics and question taxonomies.

    \item \textbf{No persistent learning model.} The system treats each session independently. It does not maintain a model of the student's recurring weaknesses, learning trajectory, or improvement over time, which limits its capacity for adaptive scaffolding.

    \item \textbf{Excerpt matching fragility.} Server-side validation now discards any excerpt that is not an exact substring of the essay, removing the risk of highlighting hallucinated quotations. The highlighting itself, however, still relies on exact substring matching between the excerpt and the editor content, so minor formatting differences (whitespace, punctuation) can leave a legitimate excerpt unhighlighted.

    \item \textbf{Non-durable study logging.} The research pipeline appends to local disk. On a serverless deployment (e.g., Vercel) the filesystem is ephemeral and effectively read-only, so durable production data capture requires moving the pipeline to a managed datastore before any longitudinal claims can be made.

    \item \textbf{No authentication or rate limiting.} The deployed endpoints are not yet behind authentication or per-user rate limiting; a public deployment that carries a server-side key needs access control and a hard spend cap to avoid being used as an open, key-bearing relay.
\end{enumerate}

\subsection{Future Work}

Building on the Build Week prototype, we identify the following directions for development and evaluation:

\subsubsection{Design-Based Research}

We plan to adopt a design-based research (DBR) methodology~\cite{noroozi2016relations} that embeds iterative product refinement within a rigorous research framework. This approach invites actual student users into the design iteration cycle, ensuring that system improvements are driven by observed learning behaviors rather than developer assumptions.

\subsubsection{IRB Approval and Empirical Evaluation}

A controlled study is planned to evaluate the system's impact on:
\begin{itemize}[leftmargin=*]
    \item Argumentative essay quality (measured via established rubrics targeting claim precision, warrant strength, counterargument depth, and evidence integration).
    \item Student cognitive engagement (measured via think-aloud protocols and, potentially, physiological markers following the methodology of~\cite{kosmyna2025brain}).
    \item Revision behavior patterns (comparing gated vs.\ ungated feedback conditions on the frequency and depth of substantive revisions).
\end{itemize}

\begin{figure}[p]
\centering
\includegraphics[height=0.9\textheight,keepaspectratio]{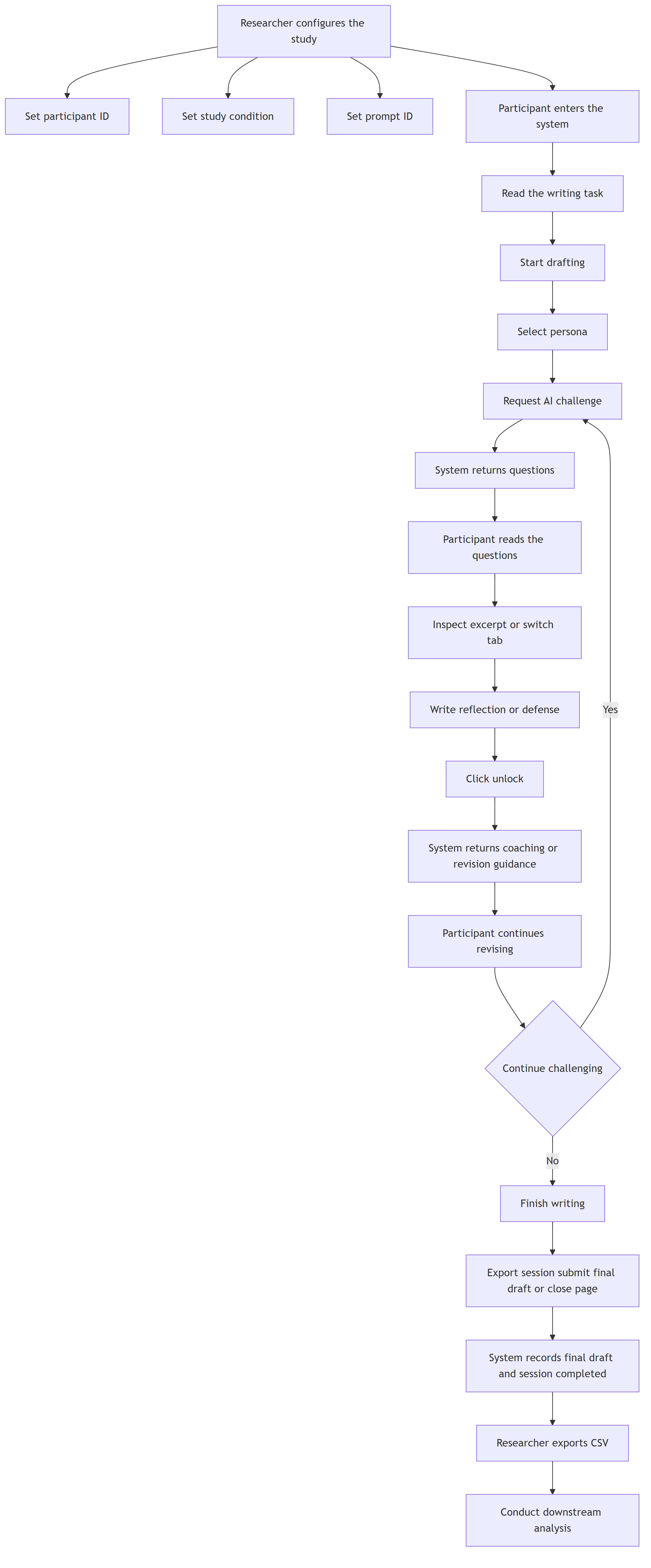}
\caption{Planned experimental workflow. The researcher configures the participant, condition, and prompt; the participant drafts, requests challenges, writes reflections, and unlocks guidance in a loop; and the system records the final draft and session-completion signals for CSV export and downstream analysis.}
\label{fig:experiment_workflow}
\end{figure}

The instrumentation described in Section~\ref{sec:instrumentation} already scaffolds such a study. Figure~\ref{fig:experiment_workflow} shows the intended experimental workflow, from researcher configuration through participant writing and the challenge/unlock loop to session export and downstream analysis.

\subsubsection{Architectural Extensions}

Technical extensions under consideration include:
\begin{itemize}[leftmargin=*]
    \item \textbf{Multi-turn defense dialogue:} Extending the reflection gate's coaching loop into an open-ended dialogue, where the tutor asks follow-up questions about an inadequate defense before releasing the suggestion, rather than the current fixed number of coached attempts.
    \item \textbf{Streaming responses:} Implementing server-sent events (SSE) for the model response to reduce perceived latency.
    \item \textbf{Student modeling:} Consuming the accumulating study logs to maintain a persistent profile of recurring weaknesses, enabling adaptive question difficulty and targeted scaffolding across sessions.
    \item \textbf{Classroom integration:} Developing a teacher dashboard that aggregates session exports across a class, identifying common argumentative weaknesses and informing instructional planning.
    \item \textbf{Additional personas:} Introducing domain-specific personas (e.g., a ``Skeptical Scientist'' for STEM argumentation, a ``Policy Analyst'' for civic discourse) to extend the system's applicability.
    \item \textbf{Durable data pipeline and production hardening:} Moving study logging to a managed datastore and placing the model-backed endpoints behind authentication, per-user rate limiting, and a spend cap---prerequisites for any at-scale classroom deployment.
\end{itemize}

\subsubsection{Collaboration and Scaling}

We are actively seeking institutional collaborations to deploy \textsc{Prober.ai} in authentic classroom settings, with the goal of collecting ecologically valid data on the system's pedagogical impact. The serverless Vercel deployment architecture already supports horizontal scaling; the primary bottleneck for production deployment is establishing rate limiting and API key management for institutional accounts.

\section{Conclusion}
\label{sec:conclusion}

\textsc{Prober.ai} demonstrates that large language models can be effectively constrained to serve as \textit{cognitive catalysts} rather than \textit{cognitive replacements} in writing education. By engineering persona-specific system prompts with explicit negative constraints, schema-constrained Structured Outputs, and a reflection-gated two-phase interaction architecture, we transform GPT-5.6 into a focused, pedagogically principled questioning engine that refuses to do the student's thinking.

The system's core architectural insight---that \textit{pedagogical friction is a feature, not a bug}---challenges the prevailing design philosophy in AI-assisted education, which overwhelmingly optimizes for reducing cognitive effort. By deliberately increasing the effort required to access revision support, \textsc{Prober.ai} ensures that students must engage in the metacognitive processes (reflection, self-explanation, argumentation defense) that produce durable learning gains.

The prototype validates the technical feasibility of this approach: a constrained GPT-5.6 reliably produces structured, inquiry-based feedback aligned with argumentation theory, and the reflection gate enforces genuine reflective engagement---coaching thin defenses rather than answering them---before delivering suggestions. Developed during OpenAI Build Week and now fully instrumented for classroom study, \textsc{Prober.ai} is positioned for the rigorous empirical evaluation through controlled classroom studies that remains the critical next step in establishing its pedagogical efficacy and informing its evolution from prototype to production educational tool.


\appendix
\section{Pedagogy Guide (Abridged)}
\label{app:pedagogy}

The following is an abridged version of the \texttt{pedagogy\_guide.md} document injected into every \texttt{/challenge} prompt. The full document (128 lines) specifies inquiry-only feedback principles, argument structure heuristics, diagnostic triggers, question module templates, and global constraints.

\begin{tcolorbox}[colback=gray!5, colframe=gray!50, title=Inquiry-Only Feedback Principles]
\small
\begin{itemize}[leftmargin=*,noitemsep]
    \item Focus on \textbf{questioning}, not correcting.
    \item Do \textbf{not} rewrite the student's text.
    \item Do \textbf{not} evaluate quality with words like ``unclear,'' ``weak,'' or ``insufficient.''
    \item Avoid yes/no questions. Ask \textbf{open-ended} questions that invite explanation.
    \item Keep cognitive load manageable: at most \textbf{3 focused questions} per module.
\end{itemize}
\end{tcolorbox}

\begin{tcolorbox}[colback=gray!5, colframe=gray!50, title=Diagnostic Triggers (Internal Only)]
\small
\begin{itemize}[leftmargin=*,noitemsep]
    \item Overgeneralization (e.g., ``all,'' ``always,'' ``never'' with limited evidence)
    \item Evidence--reason gap (facts without ``therefore'' / ``this suggests'' language)
    \item Weak counterargument (brief, strawman, or unengaged opposing view)
    \item Conceptual ambiguity (key term repeated but never defined)
    \item Causal leap (strong ``leads to'' with no mechanism)
    \item Normative claim without value framework (``should'' without stated values)
    \item Lack of stakes (no implications or ``why this matters'')
\end{itemize}
\end{tcolorbox}

\section{Sample System Prompt (Reviewer \#2)}
\label{app:prompt}

The following is the complete system prompt prefix for the Reviewer \#2 persona, illustrating the constraint methodology:

\begin{tcolorbox}[colback=gray!5, colframe=gray!50, title=Reviewer \#2 System Prompt, breakable]
\small\ttfamily
You are ``Reviewer 2'': a high-level academic peer reviewer with deep expertise.\\
Your Perspective: Expert. You assume the author should be rigorous. You are allergic to logical leaps, weak evidence, and circular reasoning.\\
Your Task:\\
1. Ignore prose, grammar, or flow. Focus strictly on the structural integrity of the argument.\\
2. Identify the single most significant logical ``black hole'' or theoretical flaw.\\
3. Pose one sharp, challenging question that forces the author to defend their core thesis.\\
Tone: Cold, clinical, and intellectually demanding. Do NOT suggest fixes. Do NOT be polite.\\
4. Ask one claim, one reasoning, one counterargument question, and one scope or implication question.
\end{tcolorbox}

\section{Research Log Schema}
\label{app:logschema}

\begin{figure}[p]
\centering
\includegraphics[height=0.92\textheight,keepaspectratio]{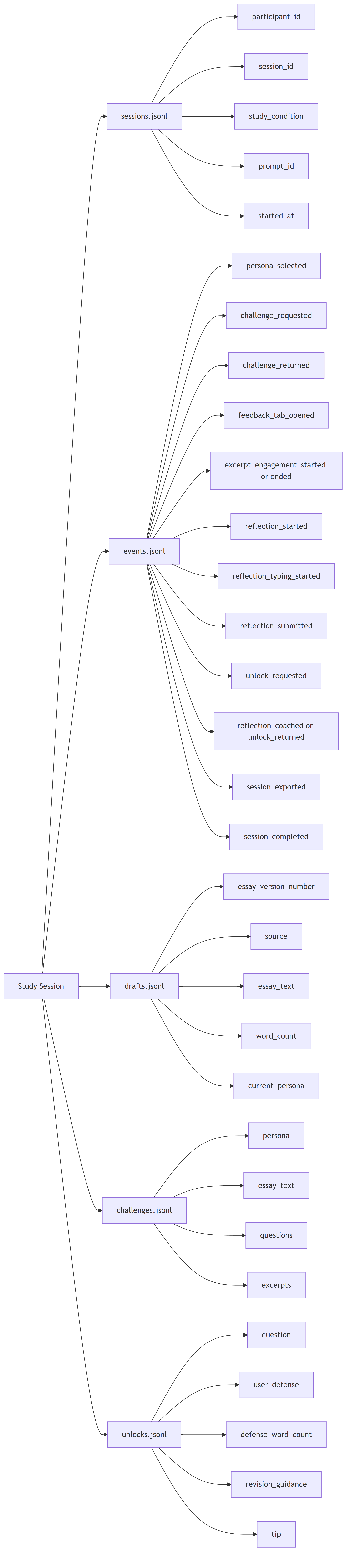}
\caption{Research data structure. Each study session produces records across five newline-delimited JSON logs; the fields shown are the columns available for downstream analysis after CSV export.}
\label{fig:data_structure}
\end{figure}

Figure~\ref{fig:data_structure} details the newline-delimited JSON logs written by the research pipeline (Section~\ref{sec:instrumentation}). Each study session fans out into five primary logs---\texttt{sessions}, \texttt{events}, \texttt{drafts}, \texttt{challenges}, and \texttt{unlocks}---whose fields together capture the participant context, the fine-grained interaction timeline, every version-numbered draft, and each question/defense/revision exchange.

\end{document}